\documentclass[letterpaper, 10 pt, conference]{ieeeconf}  %

\IEEEoverridecommandlockouts                              %

\usepackage{graphicx} 

\usepackage{amsmath,amssymb,amsfonts}
\usepackage{cite}
\usepackage{multirow}
\usepackage{algorithmic}
\usepackage[ruled,vlined,lined,linesnumbered,boxed,commentsnumbered]{algorithm2e}
\usepackage{booktabs}
\usepackage[caption=false]{subfig}
\usepackage{color}
\usepackage{balance}

\title{\LARGE \bf
Low-latency Visual SLAM with Appearance-Enhanced Local Map Building
}

\author{Yipu Zhao$^{1}$, Wenkai Ye$^{1}$, and Patricio A. Vela$^{1}$%
\thanks{$^{1}$%
Yipu Zhao, Wenkai Ye, and Patricio A. Vela  
are with School of Electrical and Computer Engineering, 
Georgia Institute of Technology, Atlanta, Georgia, USA. 
{\tt\small \{yzhao347,wye1206,pvela\}@gatech.edu}. }%
\thanks{
This work was supported in part by the China Scholarship Council 
(CSC Student No: 201606260089) and the National Science Foundation 
(Award \#1816138).
}%
}

\usepackage{url}
\usepackage{slam_macros}

\graphicspath{ {./images/}}

\begin{document}

\maketitle
\thispagestyle{empty}
\pagestyle{empty}

\begin{abstract}
A local map module is often implemented in modern VO/VSLAM systems 
to improve data association and pose estimation.  
Conventionally, the local map contents are determined by
co-visibility.  While co-visibility is cheap to establish, it 
utilizes the relatively-weak temporal prior (i.e. seen before, likely to
be seen now), therefore admitting more features into the local map than
necessary. This paper describes an enhancement to co-visibility local
map building by incorporating a strong appearance prior, which leads to 
a more compact local map and latency reduction in downstream data
association.  
The appearance prior collected from the current image influences the
local map contents: only the map features visually similar to the current 
measurements are potentially useful for data association.  
To that end, mapped features are indexed and queried with Multi-index
Hashing (MIH).  
An online hash table selection algorithm is developed to further reduce
the query overhead of MIH and the local map size.  
The proposed appearance-based local map building method is integrated
into a state-of-the-art VO/VSLAM system.  
When evaluated on two public benchmarks, the size of the local map, as
well as the latency of real-time pose tracking in VO/VSLAM are
significantly reduced.  Meanwhile, the VO/VSLAM mean performance is
preserved or improves.
\end{abstract}

\section{Introduction}

Augmentation of the feature matching process of VO/VSLAM systems with a
local map matching sub-process aids data association and state optimization
\cite{ORBSLAM,MuTa_2017_VIMapReuse}.  
Compared with a global map containing all historical 3D points, 
the local map includes only the subset of 3D points that are
hypothesized to be currently visible.  Conducting data association and
downstream state optimization on a compact local map is more efficient
than for the larger global map.

\begin{figure}[t]
  \centering
  \subfloat
    {\includegraphics[width=0.54\linewidth]{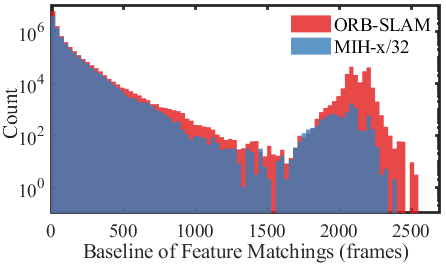}}  
  \subfloat
    {\includegraphics[width=0.44\linewidth]{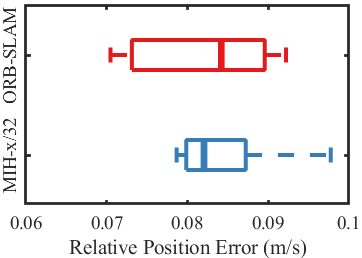}}\\
  \vspace*{-0.75em}
  \subfloat
    {\includegraphics[width=\linewidth]{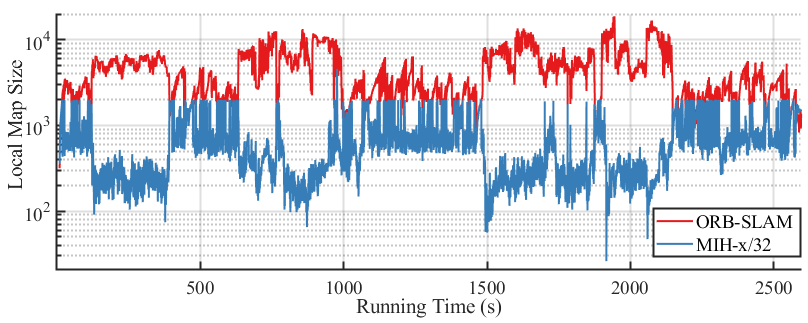}}\\
  \vspace*{-0.75em}
  \subfloat
    {\includegraphics[width=\linewidth]{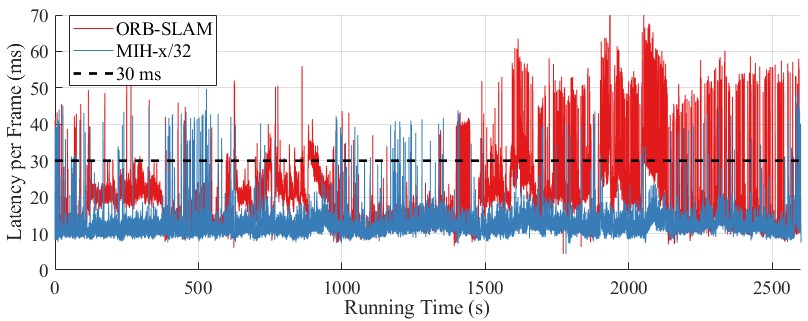}}
  \vspace*{-0.5em}
  \caption{Latency reduction of the proposed local map building
  algorithm (MIH-x/32), when integrated into a state-of-the-art VSLAM 
  system (ORB-SLAM\cite{ORBSLAM}).
  \textbf{Top-Left:} Histogram of matched features baselines 
    extracted from local map, with and without proposed algorithm
  \textbf{Top-Right:} Accuracy of VSLAM with or without proposed
    algorithm, measured with RPE (10-sec window).
  \textbf{Middle:} Size of the local map utilized in VSLAM, with or
    without proposed algorithm. 
  \textbf{Bottom:} Latency profile of real-time pose tracking on the
    long-term \textit{NewCollege} sequence.
  \label{fig:Overview}} 
  \vspace*{-1.5em}
\end{figure}

By matching 2D features from the current frame to the local map (which 
includes 3D points observed at earlier frames), extra long-baseline
feature matchings can be extracted and utilized in state optimization;
see Figure \ref{fig:Overview} (top-left) depicting a histogram of
matched local map points for ORB-SLAM, where the baseline is measured in
terms of how long ago the features were seen (as opposed to how far
spatially).
These long-baseline matchings contribute to the accuracy and
robustness of VO/VSLAM.  Not surprisingly, VO/VSLAM systems 
employing a local map \cite{leutenegger2015keyframe,ORBSLAM} tend to be more accurate and
robust than systems relying only on frame-to-frame tracking 
\cite{shen2015tightly,qin2018vins,mohta2018experiments}.

To increase the likelihood of finding and utilizing long-baseline feature 
matching, it is natural to maintain a history of the 3D points observed
earlier in time within the local map.  Specific properties or information
has been utilized to guide the local map contents to ensure a compact
yet relevant of local map, as there is a trade-off between size and
search efficiency.
The most commonly used property to guide the search of relevant 
3D points is \textit{co-visibility}.  Co-visibility was introduced for 
loop closing in VSLAM \cite{mei2010closing}, and later extended to 
pose tracking \cite{strasdat2011double,ORBSLAM,burki2016appearance,nitsche2017constrained}.  
The assumption of co-visibility being: if an earlier keyframe shares many 
3D points with a recent keyframe (i.e. co-visible), then all 3D points 
observed by the earlier keyframe are likely to be seen also.  
Co-visibility information is cheap to obtain as the by-product of
earlier data association calculations, therefore it can be considered to
be an efficient heuristic for local map building.  However, co-visibility
only utilizes the relatively-weak temporal prior (i.e. seen before,
likely to be seen now).  A local map generated with co-visibility could
easily grow without bound, and introduce significant latency to VO/VSLAM
thereafter. Figure \ref{fig:Overview} (middle row) includes a plot of
the ORB-SLAM local map versus time, where it is seen to occasionally
grow to be one to two orders of magnitude more than the number of
tracked features per frame (typically on the order of $10^2$ to $10^3$).

In this work, we propose to enhance the co-visibility local map building
step with a strong appearance prior, which will lead to a compact yet
relevant local map, a indicated in Figure \ref{fig:Overview} (middle
row) where the proposed local map queried is bounded in size and can be
up to an order of magnitude lower that for ORB-SLAM.  
The idea is straightforward: only those 3D points that are visually
similar to currently extracted features are potentially useful in data
association (and state optimization thereafter).  
To utilize the appearance prior efficiently, we propose to index
descriptors of historical 3D points with Multi-Index Hashing (MIH)
\cite{greene1994multi}.  By querying historical 3D points from a series
of hash tables, we can collect the subset of 3D points that are similar
to current measurements in appearance/descriptor space.  The
visually-similar 3D points are then verified with co-visibility, and put
together as the local map for the costly computations, e.g. data
association and state optimization.

Furthermore, an online table selection algorithm is developed to choose
a subset of hash tables that cover the most relevant 3D points.  
By only querying 3D points from the subset, the overhead on hash table
queries is reduced, while the quality of the local map is preserved,
as indicated by comparable RPE in Fig~\ref{fig:Overview} (top-right). 
The table selection process is rooted in the submodular 
property with regards to the table selection metric (e.g. information 
gain of feature matchings obtained from each table).  Because of the 
submodular property of table selection metric, a greedy algorithm can 
achieve near-optimal table selection outcomes with good efficiency
properties.  Figure \ref{fig:Overview} (bottom row) shows better
bounding of the SLAM latency per frame, with fewer outliers, relative to
a $30$ms threshold.

The proposed appearance-enhanced local map building method is integrated 
into a state-of-the-art VO/VSLAM system, ORB-SLAM \cite{ORBSLAM}.  
When evaluated on multiple public benchmarks, the size of the local map 
is significantly reduced.  More importantly, the proposed method has 
lower latency than the state-of-the-art VO/VSLAM systems, while
remaining one of the best methods in terms of accuracy and robustness.
Furthermore, the proposed local map building method is generic; it can
be easily extended to other visual(-inertial) SLAM systems 
utilizing a local map, i.e., \cite{leutenegger2015keyframe,mur2017visual}.

\section{Related Works}
This section reviews existing works that index 3D points in a map.
Two closely-related fields are explored: Vision-based Localization (VBL) 
\& Visual SLAM (VSLAM).  Differences between existing works 
and the proposed work are discussed.

VBL aims to retrieve the 6DoF pose of a visual query (image or video)
within a huge, pre-built spatial representation, e.g. a 3D point map.  
One key component of VBL is to index the spatial representation 
for efficient query.  Co-visibility was introduced to feature-based 
VBL \cite{li2010location,choudhary2012visibility} as a cue to 
prioritize feature matching efforts.  Researchers also proposed
alternative indexing methods based on appearance/feature descriptors
\cite{sattler2011fast,lim2012real}.  
Real-valued feature descriptors such as SIFT\cite{SIFT} and SURF
\cite{SURF} are typically indexed offline using a kd-tree.  
Appearance-based indexing are proven to yield more accurate \& 
robust query results, while co-visibility is more computationally-efficient.  
Combining both cues was first explored in \cite{sattler2012improving}, 
and further refined in \cite{sattler2017efficient,lynen2015get}.
The work \cite{lynen2015get} replaced the kd-tree data structure with a 
faster \& more flexible indexing method, inverted multi-index.  
The appearance-based query results are then filtered with co-visibility.
Such a combination scheme is efficient: the VBL system runs real-time on
mobile device.  Nevertheless, training the inverted index is still an
offline process requiring a known 3D map.  

Recently, binary feature descriptors such as BRISK \cite{leutenegger2011brisk} and 
ORB \cite{rublee2011orb} have become popular in VBL since they are more efficient
to extract.  Conventional indexing data structures like kd-trees are
better suited to real-valued descriptors, rather than binary ones,
motivating the exploration of alternative indexing methods.  
For example, \cite{feng2016fast} proposed to index binary descriptors
with randomized trees, which were trained offline from the pre-built 3D
map.  Hashing has been proven to be a good indexing solution 
\cite{cheng2014fast,tran2019device} in binary-descriptor VBL.  
Coarse-to-fine searching schemes are commonly applied in these VBL
systems, where an initial hashing query provides the coarse results that
are later refined by a linear scan.  

Apart from compatibility with binary descriptors, two other properties
of hashing make it particularly attractive to online \& incremental pose
estimation problem, e.g. VSLAM.  First, hashing index can be updated
efficiently for online processes.  
It is then possible to generate a more compact and relevant index 
by updating hash tables, e.g., according to changes in the map \& 
the visibility constraints.  Second, hashing relaxes the requirement 
for database pre-training (or prior offline database generation),
therefore enabling VSLAM systems to operate in general and unknown
environments.  
Hashing has been applied to modules of VSLAM where real-time performance
is not required.  \cite{straub2013fast} indexed binary descriptors 
with Locality Sensitive Hashing (LSH) \cite{LSH}, and demonstrated 
good relocalization performance in a VSLAM system.  
\cite{han2017mild} utilized Multi-Index Hashing (MIH)
\cite{greene1994multi} in the loop closing module of VSLAM.  

The proposed work is based on MIH, but with a key enhancement: 
an online table selection algorithm is developed to reduce the 
number of hashing queries, therefore enabling MIH to be used in VSLAM 
modules with real-time requirements, e.g. pose tracking. 
The local map queried with appearance/feature descriptors is further 
tailored with a co-visibility check.  
The final local map is more compact than the ones generated with either
co-visibility or appearance only.  Running data association and state
optimization on the size-reduced local map is more efficient and 
leads to significant latency reductions in VSLAM based on a more
efficient local map data association step.  
Furthermore, the quality of the local map (e.g. amount of long-baseline
feature matchings) is preserved in the compact local map.  Therefore,
the performance of VSLAM is preserved. Preliminary quantification of
these benefits can be seen in Figure \ref{fig:Overview} for a single
sequence.

\section{Local Map Building with Multi-Index Hashing}
\begin{figure}[!tb]
  \centering
  \includegraphics[clip, trim=0cm 13.7cm 2cm 0.5cm, width=\linewidth]{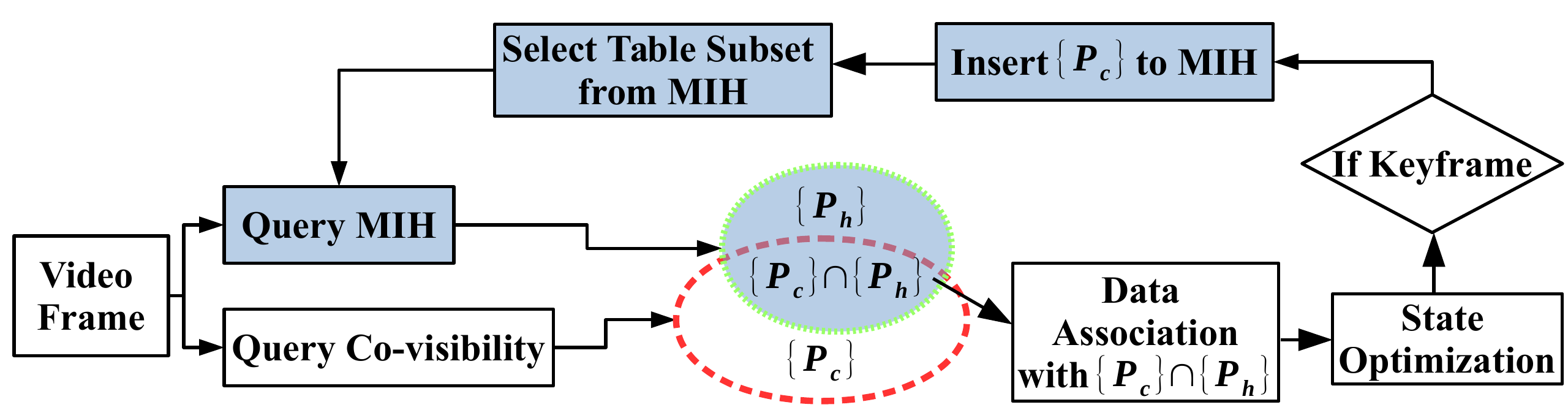}  
  \caption{Framework of the proposed local map building method.  The 
  local map built with co-visibility is the red dashed ellipse, while 
  the one built by querying MIH is the green dashed ellipse.  
  Their intersection defines the local map for downstream
  processing, i.e. data association and state optimization. 
  \label{fig:Framework}} 	
  \vspace*{-1em}
\end{figure}

A diagram of the proposed local map building method is illustrated
in Fig~\ref{fig:Framework}.  The modules of proposed method are
highlighted with shaded boxes, while those in a conventional VSLAM
pipeline have clear boxes.  This section describes the query and
insertion stage of MIH.  The hash table selection algorithm will be
introduced in the next section.

\vspace*{.25em}
\noindent
\textbf{Query MIH.}
Assume that a frame with $m$ binary descriptors extracted is provided 
and that the MIH contains $t$ hash tables.  
Each binary descriptor will trigger a MIH query.
In a MIH query, the $b$-bit binary query descriptor is first 
separated into $t$ disjoint contiguous substrings.  
Each substring gets queried with the corresponding hash table for an
exact match.  Query results from all $t$ hash tables are put together as
the final query result.  Repeating the MIH query for all binary
descriptors from the input frame, aggregate the 3D point set $\{ P_h \}$
that satisfy the appearance prior.  Its intersecting with the 3D point
set $\{ P_c \}$ collected with conventional co-visibility builds the final
local map $\{ P_c \} \cap \{ P_c \}$.

\vspace*{.25em}
\noindent
\textbf{Insert to MIH.}
Updating MIH according to changes in the map \& visibility constraints
is essential for efficient local map building.  
As a trade-off between update frequency and computation cost, MIH
updates are triggered only for keyframes sent to the mapping thread.  
Updating MIH in the mapping thread avoids introducing overhead during
real-time pose tracking.  

For each keyframe, the co-visible 3D points $\{ P_c \}$ are inserted into 
the MIH.  Similar to the query process, the $b$-bit binary descriptor of
each 3D point in $\{ P_c \}$ is separated into $t$ disjoint contiguous
substrings, each of which is of length $\lfloor b/t \rfloor$.  
Each substring is then inserted into a corresponding hash table.  
For 3D points already in the hash tables, their entries will be brought
to the front of the bucket, making them more likely to be queried in the
future.  

\vspace*{.25em}
\noindent
\textbf{Choice of hash table number.}
The amount of hash tables $t$ has strong impact on the
performance-efficiency of MIH-based local map indexing.  
Recall the example of a frame with $m$ features extracted.  
Each feature will trigger a MIH query consisting of $t$ queries to hash tables.
Therefore, the MIH-based local map building has 
a time complexity of $\mathcal{O}(mt)$, i.e., linear in $t$.
Meanwhile, the space complexity of MIH is 
$\mathcal{O}(tN2^{\lfloor b/t \rfloor})$, 
where $N$ is the bucket size in each hash table.  
The space complexity decreases exponentially with table number $t$.  
Therefore, only a certain range of $t$ works in practical applications 
due to time \& space complexity limits.

Apart from time \& space complexity, the robustness of local map 
building against perturbations in binary descriptors is largely 
decided by hash table number $t$.  
Assuming $\epsilon$ bits of the query descriptor are perturbed 
under a uniform distribution, the recall probability (i.e. probability 
that the query succeeds with a perturbed string) is 
connected to hash table number $t$ as per \cite{han2017mild}:
\begin{equation} \label{eq:Recall_Prob}
  P_{recall}(t,\epsilon)=1 - t!\;\Theta(\epsilon,t)/t^\epsilon,
\end{equation}
where $\Theta(\epsilon,t)$ is the Stirling partition number 
\cite{graham1989concrete}.  

\begin{figure}[t]
  \centering
  \includegraphics[width=0.7\linewidth]{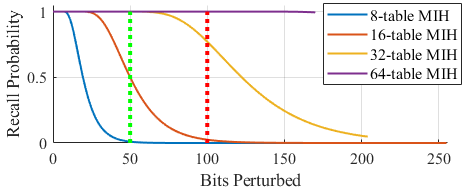}
  \vspace*{-0.5em}
  \caption{Simulation results evaluating 
  the recall probability of hashing (the higher the better)
  vs. the number of bits perturbed for different numbers of
  tables in the MIH.  For 256-bit descriptors, MIH with 32 tables is
  preferred: it remains high recall even under significant perturbation
  (50-100 bits).
  \label{fig:Simu_MIH_Table}} 	
  \vspace*{-0.75em}
\end{figure}

When working with 256-bit binary descriptors such as ORB, the
relationship described in Eq~\ref{eq:Recall_Prob} is illustrated in 
Fig~\ref{fig:Simu_MIH_Table}.
The green and red dashed lines indicate example thresholds of bit-wise 
perturbations in typical SLAM applications.  
At least 32 tables are needed for high recall probability within the
example perturbation levels (vertical dashed lines).  
Using 64 tables is also possible, but with the drawback of higher
overhead due to the linear-growth in time complexity.  
In the proposed local map indexing method, 32 hash tables are
maintained; each table covers an 8-bit descriptor substring.  

\vspace*{.25em}
\noindent
\textbf{Choice of bucket size.}
Another parameter affecting the performance-efficiency of MIH-based
local map building is the bucket size $N$ of each hash table.  
A bucket in MIH is implemented as ring buffer, where only the $N$ most
recent 3D points are stored.  
For the purpose of long-baseline feature matching, it is necessary to
keep the entries of 3D points observed earlier in time within the bucket.  
However, an over-sized bucket will store entries of 3D points that are no
longer visible nor relevant.  As a consequence, the resulting local map
will be less compact and relevant, introducing overhead to data association.  
In what follows, the bucket size $N$ is set to 10 based on a parameter sweep.

\section{Overhead Reduction with Hash Table Selection}
For a frame with $m$ features extracted and a 32-table MIH,  
the number of hash table queries in local map building is
$\mathcal{O}(32m)$.  While querying all 32 hash tables 
provides robustness against severe perturbation, querying 
a subset of hash tables is more efficient when the bit-wise perturbation 
level is low or medium.  We propose an online table selection algorithm
to identify the minimum subset of hash tables to be queried, which
further improve the compactness of local map without performance
degeneration.

\vspace*{0.25em}
\noindent
\textbf{Formulation.}
To begin, the metric used for table selection is introduced. 
Assume $F$ is the full set of {\em true} feature matchings between 
current frame and the full local map built with all 32 hash tables.  
For each hash table $T_i$, the {\em true} feature matchings that 
can be queried from it form a subset $F_i \subset F$, where  
$\bigcup_{i=1}^{32}{F_i} = F$.  For each hash table $T_i$, the
contribution towards current state optimization can be assessed with the
information matrix of subset $F_i$.  

The least squares objective of VO/VSLAM pose tracking is
\begin{equation}  \label{eq:LeastSquare}
  \min \left\Vert h(x, p)-z \right\Vert^2,
\end{equation}  
where $x$ is the pose of the camera, 
$p$ are the 3D feature points and 
$z$ are the corresponding 2D image measurements.  
The measurement function, $h(x, p)$, is a combination of the $SE(3)$
transformation (world-to-camera) and pin-hole projection.
To first-order approximation, the information matrix of the camera 
pose $\Omega_x$ is
\begin{equation} \label{eq:Pose_InfoMat}
  \Omega_x = \sum H(i)^T \Omega_r(i) H(i) = \sum \Omega_x(i),
\end{equation}
where $H(i)$ and $\Omega_r(i)$ are the measurement Jacobian and 
residual information matrix of corresponding {\em true} matched features.  
Denote by $\Omega_x(i)$ the pose information matrix derived from a
single feature match $i$.

As introduced for feature subset selection 
\cite{carlone2019attention,zhao2018good2}, the {\em logDet} is
especially suited for quantifying the contribution of matched features 
to VO/VSLAM.  Therefore, the value of a hash table $T_i$ towards current
state optimization can be measured with 
\begin{equation} \label{eq:Table_Metric}
  \log \det (\sum_{i\in F_i} \Omega_x(i)).
\end{equation}

There is a certain level of overlap between the {\em true} matched
feature subsets for each hash table.  
In ideal scenario without any perturbation to feature descriptor, 
the full set of {\em true} feature matchings can be retrieved from 
any one of the 32 hash tables, i.e. 100\% overlapping between subsets, 
$\forall i,j\ F_i = F_j = F$.  In practice perturbations reduce the
subset overlap percentage to less than 100\%, and each hash table covers
a subset of {\em true} feature matchings $F$.  Therefore, selecting a
subset of hash table is equivalent to a problem of maximum coverage,
with the objective formulated as:
\begin{equation} \label{eq:Selection_Obj}
  \max_{S\subseteq\{1,2,...,32\}, |S|\leq k} \log \det(\sum_{i\in \{\bigcup_{h\in S}{F_h}\}} \Omega_x(i)), 
\end{equation}
where $k$ is the cardinality constraint.  

\vspace*{0.25em}
\noindent
\textbf{Greedy Solution.}
The maximum coverage problem is studied in the field of computational
theory, where it is known to have submodular properties. Of note,
\begin{theorem} \cite{goemans1995minimizing}
\label{theo:Selection_Greedy}
Let $f$ be a monotone submoduar function, then greedy algorithms
achieve a $\left( 1-1/e \right)$ approximation guarantee 
to the optimum solution of Eq~\eqref{eq:Selection_Obj}.
\end{theorem} 

As proven in \cite{shamaiah2010greedy}, {\em logDet} is submodular \&
monotone increasing.  
Solutions to the subset selection problem, and the equivalent hash table
selection problem, can be approximated using greedy algorithms.
More importantly, a greedy algorithm is guaranteed to be near-optimal,
with approximation ratio of $1-1/\epsilon$.  Based on this outcome,
we present a greedy, online hash table selection algorithm in 
Alg~\ref{alg:tableSelection}.  Two control parameters are fixed after 
parameter sweep: cardinality constraint $k=8$, target contribution 
$d_{thres}=80.0$.

\begin{algorithm}[t]
 \KwData{$\text{feature matching subset from each hash table}\ $
 $ \{F_1,\ F_2,\ ...\ ,\ F_{32}\}$,\ $\text{cardinality constraint}\ k$,\ $\text{target contribution}\ d_{thres}$}
 \KwResult{ $\text{indices of hash tables selected}\ S$}

 \ForEach{$\text{feature matching}\ j\in \bigcup_{i=1}^{32}{F_i}$}{			%
	 $\text{collect pose information matrix}\ \Omega_x(j)$\;
 }

 $S\gets \emptyset$,\ $d_{acc}=0$;
 
 \While{$|S|<k \land d_{acc}<d_{thres}$}{								%

	\ForEach{$i \notin S$}{
		$d(i) = \log \det(\sum_{i\in \{\bigcup_{h\in S \cup F_i}{F_h}\}} \Omega_x(i))$
	} 

 	$j \gets \arg\max_{i} d(i)$;
 	
 	$d_{acc}=d(j)$;
 	
	$S \gets S \cup j$;

 }
 \Return $S$.
 \caption{Online hash table selection algorithm. \label{alg:tableSelection}}
\end{algorithm}

Notice that the above discussion assumes that the {\em true} feature 
matchings are known whem performing hash table selection.  
We assume that the hash table contents are a slowly-varying function of
time.  Therefore, the hash table subset selection algorithm runs at a
lower rate than real-time pose tracking, and only updates the selected
subsets at keyframes.  Between keyframes, the hash table subset
queried for local map building is fixed.

\section{Experimental Results}
This section evaluates the performance-efficiency trade off of the 
proposed local map building algorithm on a state-of-the-art VSLAM 
system, ORB-SLAM\cite{ORBSLAM}.  Applying the proposed algorithm 
to the real-time tracking thread of ORB-SLAM reduces pose tracking 
latency.  Meanwhile, tracking accuracy is either improved (on short
sequences) or remains near the same level as canonical ORB-SLAM 
(on long sequence), and the robustness is preserved 
(i.e. avoid tracking failure).
 
Two public benchmarks are used to evaluate the proposed algorithm:
\setlength{\IEEEelabelindent}{0em}
\begin{enumerate}
  \item\textit{NewCollege} \cite{smith2009new}, which 
    contains a 43-minutes stereo sequence collected with 
    a robot traversing a campus and adjacent parks.  There 
    are multiple loops/revisits within the sequence.  The sequence 
    is well-suited for evaluating the long-term performance 
    \& efficiency of VSLAM system (with loop closure).  
    Due to the lack of 6DoF pose ground truth, offline 
    Bundle Adjustment is executed with stereo video, and 
    the jointly optimized camera poses are taken as the ground truth.  
    We only evaluate monocular VSLAM (e.g. with left camera) 
    against the ground truth in this experiment.
 \item \textit{EuRoC} \cite{burri2016euroc}, which contains 
    11 stereo-inertial sequences comprising 19 minutes of video,
    recorded in 3 different indoor environments.  Compared with 
    \textit{NewCollege}, videos in \textit{EuRoC} are well-suited
    for evaluating the short-term performance \& efficiency 
    of VO (without loop closure).  Ground-truth tracks are provided
    using motion capture systems (Vicon \& Leica MS50).  
    We evaluate only monocular VO implementations on \textit{EuRoC}.
\end{enumerate}

Two performance metrics are used in the experiment.  When evaluating 
the short-term performance of VO on \textit{EuRoC}, 
\textit{absolute root-mean-square error} (RMSE) between ground truth 
track and real-time VO estimation is used.  When evaluating the long-term
performance of VSLAM on \textit{NewCollege}, the Relative Position Error (RPE) 
\cite{sturm12iros_ws,KITTI} is chosen.  Compared with absolute RMSE, 
RPE is less sensitive to the inevitable scale drift of monocular VSLAM.  
Therefore, it is better for evaluating monocular systems on long-term 
sequences.  

The efficiency of VO/VSLAM is evaluated with the latency of real-time
pose tracking per frame, which is defined as the time interval from
receiving an image to publishing the pose estimate.  
Latency of mapping \& loop closing is less of a concern in this work due
to the relaxed time constraints of those processes.

Performance assessment involves a 10-run repeat for each configuration,
i.e., the benchmark sequence, the VO/VSLAM approach and the parameter 
(number of features tracked per frame).  
Results for a tested VO/VSLAM configuration are discarded if at least 
one run experiences track loss.
The experiments are conducted on a desktop equipped with an Intel i7
quadcore 4.20GHz CPU (passmark score of 2583 per thread) running the ROS
Indigo environment.  

\begin{figure}[!t]
  \centering
  \vspace*{0.25em}
  \includegraphics[clip, trim=3.5cm 7cm 4cm 8cm, width=\linewidth]{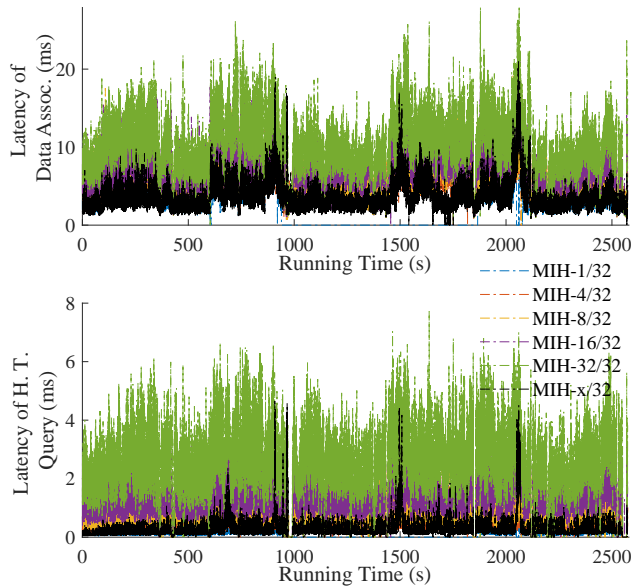}  
  \vspace*{-1.5em}
  \caption{
    \textbf{Top:} Latency of data association from 1 run on 
      \textit{NewCollege}.
	\textbf{Bottom:} Latency of hash table query (part of data association) 
      from 1 run on \textit{NewCollege}.
	The first 5 profiles have predefined hash table subsets, e.g. first 1, 
      first 4, etc.  The last profile employs online hash table subset
      selection.
  \label{fig:LatencyProf_OTS}} 	
  \vspace*{-0.15in}
\end{figure}

\subsection{Online Table Selection vs Fixed Table Subsets}
To demonstrate the benefit of online hash table selection 
(Alg~\ref{alg:tableSelection}), we performed additional 10-run repeats
of MIH-based local map building with a predefined set of fixed hash
table subsets, ranging 1 table (MIH-1/32) to all 32 tables (MIH-32/32).
Results of these tests are compared to MIH-based local map building with
online hash table selection, i.e. MIH-x/32 (x = 10).

\begin{figure}[!t]
  \centering
  \includegraphics[width=\linewidth]{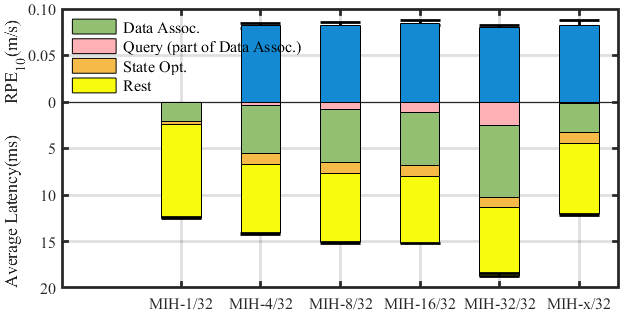} 
  \caption{
	RPE \& latency for different hash table subsets averaged over 
    10 runs on \textit{NewCollege}.  The first 5 columns are the fixed hash 
    table subset methods, e.g. first 1, first 4, etc.
	The last column employs online selection.
	No RPE is reported for the single hash table (MIH-1/32) since track
    loss frequently occurred.
    \label{fig:RPELatency_OTS}} 	
  \vspace*{-0.15in}
\end{figure}

The latency profiles of different hash table subsets are presented 
in Fig~\ref{fig:LatencyProf_OTS}.  MIH-x/32 has the lowest latency for
data association, when compared to other predefined hash table subsets.
The latency of hash table queries is also lower with online hash table
selection.  Performance evaluation of the methods collected the
average RPE (with a 10-sec window), and also logged the average latency
of each module in the real-time pose tracking process. Performance (RPE) and 
efficiency (latency) outcomes are summarized in Fig~\ref{fig:RPELatency_OTS}.
MIH-x/32 has the lowest latency for pose tracking while preserving the
performance of VSLAM relative to the fixed table subsets.

\subsection{Comparison with State-of-the-Art VO/VSLAM}
The latency reduction and strong performance of the proposed local map
building algorithm is demonstrated by comparing with other
state-of-the-art VO/VSLAM systems.  

\vspace*{0.25em}
\noindent
\textbf{VSLAM.}
Two state-of-the-art VSLAM systems are chosen as baselines: DSO with 
loop closure (LDSO) \cite{gao2018ldso} and ORB-SLAM (ORB) \cite{ORBSLAM}.  
In addition to the proposed MIH-x/32, we integrate two reference 
methods into ORB-SLAM that enhance co-visibility local map building 
with simple heuristics.  One heuristic is random sampling, i.e. {\em Rnd}.  
The other heuristic prioritizes map points with a long track history,
denoted as {\em Long}, since feature points tracked for a long time are
more likely to be mapped accurately.  

\begin{figure}[!tb]
	\centering
	{\includegraphics[width=\linewidth,clip=true,trim=0.10in 0in 0.4in
    0in]{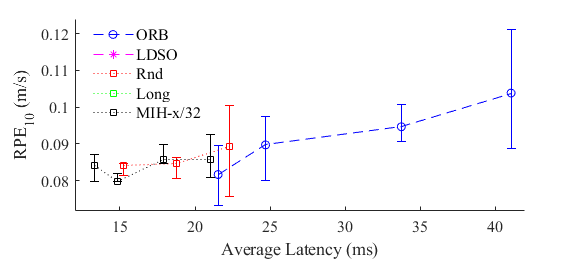}}
	\caption{Latency vs. accuracy on NewCollege monocular sequence. 
	System evaluation involved a sweep of features per frame: 
    800, 1000, 1500, 2000.  
	\label{fig:NC_Time_vs_RPE}}
\end{figure}

To capture the performance-efficiency trade off of VSLAM systems, we
adjust the number of features/patches extracted per frame.  All 5 VSLAM
systems are configured to run 10-repeats on \textit{NewCollege}, with
feature/patch quantities ranging from 800 to 2000.  The RPE under 10-sec
window versus the average latency per frame is depicted in 
Fig~\ref{fig:NC_Time_vs_RPE}.
Relative to ORB-SLAM, the proposed MIH-x/32 leads to latency reduction
for all configurations of feature number.  {\em Rnd} also leads to
latency reduction, but not as much as  MIH-x/32. The {\em Rnd} case with
800 features leads to track loss, so it is not plotted.
Both LDSO and {\em Long} failed to track the full \textit{New College} 
sequence.  The accuracy of MIH-x/32 is comparable to the best performing
ORB realizations, but with a lower deviation as indicated by the shorter
error bars.
Lastly, we report the accuracy \& latency of the monocular VSLAM systems
under the configuration of 800 features per frame in Table~\ref{tab:NC_VSLAM}.
Three RPE metrics are computed using different sliding windows: 3-sec, 10-sec 
and 30-sec.  
In addition to the average RPE over 10-run repeat, the standard deviation (STD) 
of the RPE is also reported in each cell of Table~\ref{tab:NC_VSLAM}.  
The two heuristics {\em Rnd} and {\em Long} are excluded since they both
failed to track on the full sequence.  The best numbers (lowest
average/STD of RPE, lowest latency) are highlighted with bold.  
The accuracy of MIH-x/32 remains at similar levels as ORB (equal or
around 10\%), as evaluated on all 3 RPE metrics.  
More importantly, the latency of proposed method is lower and more
consistent than baseline ORB. It is 21\%, 33\%, and 40\% lower for the
first quartile, average, and third quartile values.
 
\setlength{\tabcolsep}{0.2em}
\begin{table}[t!]
	\small
	\centering
	\caption{RPE (m/s) and Latency (ms) on NewCollege Sequence}
	\centering
	\begin{tabular}{c|c|c|c|c|}
		\toprule[1.5pt]\midrule
		\multicolumn{5}{c}{\textbf{VSLAM (with loop-closure)}}\\
		\midrule
		\  \parbox[t]{2mm}{\multirow{4}{*}{\rotatebox[origin=c]{90}{\textbf{RPE (STD)}}}} \ 
		& \textbf{\ \small Seq.\ } & \ LDSO\ & \ ORB\ &\ MIH-x/32\ \\
		\cmidrule{2-5}
		& $\text{RPE}_{3}$		&	- 	&	\textbf{0.11}\ (2e-2) 	&  0.12\ \textbf{(8e-3)} \\
		& $\text{RPE}_{10}$  	&	-	&	\textbf{0.08}\ (8e-3) 	& \textbf{0.08\ (6e-3)} \\
		& $\text{RPE}_{30}$  	&	-	&	\textbf{0.09\ (5e-3)} 	& 0.10\ (1e-2) \\
 		\midrule
		\parbox[t]{2mm}{\multirow{3}{*}{\rotatebox[origin=c]{90}{\textbf{Latency}}}}
 		&\small $\boldsymbol{Q_1}$   	& - & 13.2 & \textbf{10.4} \\
		&\textbf{\small Avg.}     		& -	& 18.3 & \textbf{12.2} \\
		&\small $\boldsymbol{Q_3}$     	& -	& 21.5 & \textbf{13.3} \\
		\midrule
		\bottomrule[1.5pt]
	\end{tabular} 
	\label{tab:NC_VSLAM}
\end{table}

\vspace*{0.25em}
\noindent
\textbf{VO.}
Two state-of-the-art VO baselines are included: SVO\cite{SVO2017} 
and DSO \cite{DSO2017}.  For fair comparison, the loop closing 
module is disabled on all ORB-SLAM variants: canonical ORB, 
MIH-x/32, {\em Rnd}, and {\em Long}.  All VO systems are configured to
run 10-repeats on \textit{EuRoC} under example configuration (800
features per frame).  The short-term performance of VO are evaluated
with RMSE, while the efficiency is still assessed via per frame tracking
latency.  Accuracy \& latency results are summarized 
in Table~\ref{tab:EuRoC_VO}. 
The best value (lowest RMSE, lowest latency) in each row is highlighted 
with bold in Table~\ref{tab:EuRoC_VO}.  
According to the upper part of Table~\ref{tab:EuRoC_VO}, DSO and 
the 2 local map building heuristics are not robust enough (e.g. frequent 
track loss).  SVO tracks 9 of 11 sequences, but with the highest 
RMSE over all VO systems.  Both ORB baseline and proposed MIH-x/32 
track 8 of 11 sequences.  Additionally, MIH-x/32 improves the accuracy
relative to baseline ORB, with an RMSE average that is 41\% lower.

The latency reduction of MIH-x/32 is less significant for these
short-term VO sequences, when compared with the previous long-term VSLAM
outcomes.  Nevertheless, MIH-x/32 has the 2nd lowest average latency
among all 6 VO systems, second to SVO.  When comparing the 3rd quantile
of latency, MIH-x/32 is lower than SVO (by 3\%), which suggests that
tighter latency bounds can be achieved with the proposed local map
building algorithm.

\setlength{\tabcolsep}{0.2em}
\begin{table}[t!]
	\small
	\centering
	\caption{RMSE (m) and Latency (ms) on EuRoC Sequences}
	\centering
	\begin{tabular}{c|c|c|c|c|c|c|c|}
		\toprule[1.5pt]\midrule
		\multicolumn{8}{c}{\textbf{VO (without loop-closure)}}\\
		\midrule
		& \textbf{\small Seq.} & SVO & DSO & ORB & MIH-x/32 & Rnd & Long \\
		\midrule
				\parbox[t]{2mm}{\multirow{14}{*}{\rotatebox[origin=c]{90}{\textbf{RMSE}}}}
		&\textit{MH 01 easy}  &	0.227 & 0.407 & 0.027 & 0.026 & \textbf{0.025} & - \\ 
		&\textit{MH 02 easy}  & 0.761 & - &	0.034 & \textbf{0.031} & 0.034 & - \\ 
		&\textit{MH 03 med}   & 0.798 & 0.751 &	0.041 & 0.086 & \textbf{0.035} & - \\ 
		&\textit{MH 04 diff}  & 4.757 & - & 0.699 & \textbf{0.293} & 0.746 & 0.329 \\ 
		&\textit{MH 05 diff}  & 3.505 & - & 0.346 & \textbf{0.197} & - & - \\ 
		&\textit{VR1 01 easy} & 0.726 & 0.950 &	0.057 & 0.040 & \textbf{0.034} & - \\ 
		&\textit{VR1 02 med}  & 0.808 & \textbf{0.536} &	- & - & - & - \\ 
		&\textit{VR1 03 diff} & - & - &	- & - & - & - \\ 
		&\textit{VR2 01 easy} & 0.277 & 0.297 &	0.025 & 0.032 & \textbf{0.021} & - \\ 
		&\textit{VR2 02 med}  & 0.722 & 0.880 &	0.053 & \textbf{0.035} & 0.216 & - \\ 
		&\textit{VR2 03 diff} & - & - &	- & - & - & - \\ 
		\cmidrule{2-8}
		&\textbf{\small Avg.} & 1.477 & 0.637 &	0.160 & \textbf{0.093} & 0.159 & 0.329 \\
 		\midrule[1pt]
		\parbox[t]{2mm}{\multirow{3}{*}{\rotatebox[origin=c]{90}{\textbf{Latency}}}}
 		&\small $\boldsymbol{Q_1}$    &	7.4	& \textbf{5.8} & 13.9 & 11.4	& 12.0 & 11.3 \\ 
		&\textbf{\small Avg.}      	  &	 \textbf{12.6} & 16.4 & 18.4 & 15.7	& 16.0 & 17.7 \\ 
		&\small $\boldsymbol{Q_3}$    &	16.8 & 19.1 & 20.7 & 16.3 & \textbf{16.1} & 21.0 \\
		\midrule
		\bottomrule[1.5pt]
	\end{tabular} 
	\label{tab:EuRoC_VO}
\end{table}

\section{Conclusion}
This paper demonstrated how an appearance prior can be exploited to
build a compact yet relevant local map in VSLAM.  Working with the
compact local map leads to latency reduction in time-sensitive VSLAM
modules, i.e., pose tracking.  
Meanwhile, the accuracy and robustness of VSLAM is preserved, thanks to
the preservation of long-baseline feature associations in the local map.  
On both long-term VSLAM and short-term VO applications, the proposed
algorithm leads to significant latency reduction in real-time pose
tracking, while keeping (if not improving) VO/VSLAM performance relative
to the baseline variant and having the best performance relative to
other state-of-the-art systems.

\addtolength{\textheight}{-2cm}       %

\balance 
\bibliographystyle{IEEEtran}
\bibliography{../../full_references}

\end{document}